\pdfoutput=1

\documentclass[11pt]{article}

\usepackage{acl}

\usepackage{times}
\usepackage{latexsym}

\usepackage[T1]{fontenc}

\usepackage[utf8]{inputenc}

\usepackage{microtype}

\usepackage{colortbl}
\usepackage{graphicx}
\usepackage{subfigure}
\definecolor{mygray}{gray}{.9}
\usepackage{algorithm}
\usepackage{algorithmic}
\usepackage{color}
\usepackage{multirow}
\usepackage{amsfonts,amssymb}
\usepackage{amsmath}
\usepackage{booktabs}
\usepackage{mathtools}
\usepackage{url}
\usepackage{stfloats}
\title{Template-free Prompt Tuning for Few-shot NER}

\author{
    Ruotian Ma\textsuperscript{\rm 1}\thanks{\ \ Equal contribution.} , Xin Zhou\textsuperscript{\rm 1}\footnotemark[1] , Tao Gui\textsuperscript{\rm 2}\thanks{\ \  Corresponding authors.} , Yiding Tan\textsuperscript{\rm 1},\\
    \textbf{Linyang Li\textsuperscript{\rm 1}, Qi Zhang\textsuperscript{\rm 1}\footnotemark[2] , Xuanjing Huang\textsuperscript{\rm 1}} \\
  $^1$School of Computer Science, Fudan University, Shanghai, China \\
  $^2$Institute of Modern Languages and Linguistics, Fudan University, Shanghai, China \\
  \texttt{\{rtma19,xzhou20,tgui,qz,xjhuang\}@fudan.edu.cn}
}

\begin{document}
\maketitle
\begin{abstract}

Prompt-based methods have been successfully applied in sentence-level few-shot learning tasks, mostly owing to the sophisticated design of templates and label words. 
However, when applied to token-level labeling tasks such as NER, it would be time-consuming to enumerate the template queries over all potential entity spans. 
In this work, we propose a more elegant method to reformulate NER tasks as LM problems without any templates.
Specifically, we discard the template construction process while maintaining the word prediction paradigm of pre-training models to predict a class-related pivot word (or label word) at the entity position. Meanwhile, we also explore principled ways to automatically search for appropriate label words that the pre-trained models can easily adapt to. 
While avoiding the complicated template-based process, the proposed LM objective also reduces the gap between different objectives used in pre-training and fine-tuning,
thus it can better benefit the few-shot performance.
Experimental results demonstrate the effectiveness of the proposed method over bert-tagger and template-based method under few-shot settings. Moreover, the decoding speed of the proposed method is up to 1930.12 times faster than the template-based method.

\end{abstract}

\section{Introduction}

Pre-trained language models (LMs) have led to large improvements in NLP tasks \cite{devlin-etal-2019-bert,liu2019roberta,lewis-etal-2020-bart}. Popular practice to perform downstream classification tasks is to replace the pretrained model's output layer with a classifier head and fine-tune it using a task-specific objective function. Recently, a new paradigm, prompt-based learning, has achieved great success on few-shot classification tasks by reformulating classification tasks as cloze questions.
Typically, for each input [X], a template is used to convert [X] into an unfilled text (e.g., ``[X] It was \_\_."), allowing the model to fill in the blank with its language modeling ability. For instance, when performing sentiment classification task, the input ``I love the milk." can be converted into ``I love the milk. It was \_\_.". Consequently, the LM may predict a label word ``great", indicating that the input belongs to a positive class.

\begin{figure}
    \centering
    \includegraphics[width=0.9\linewidth]{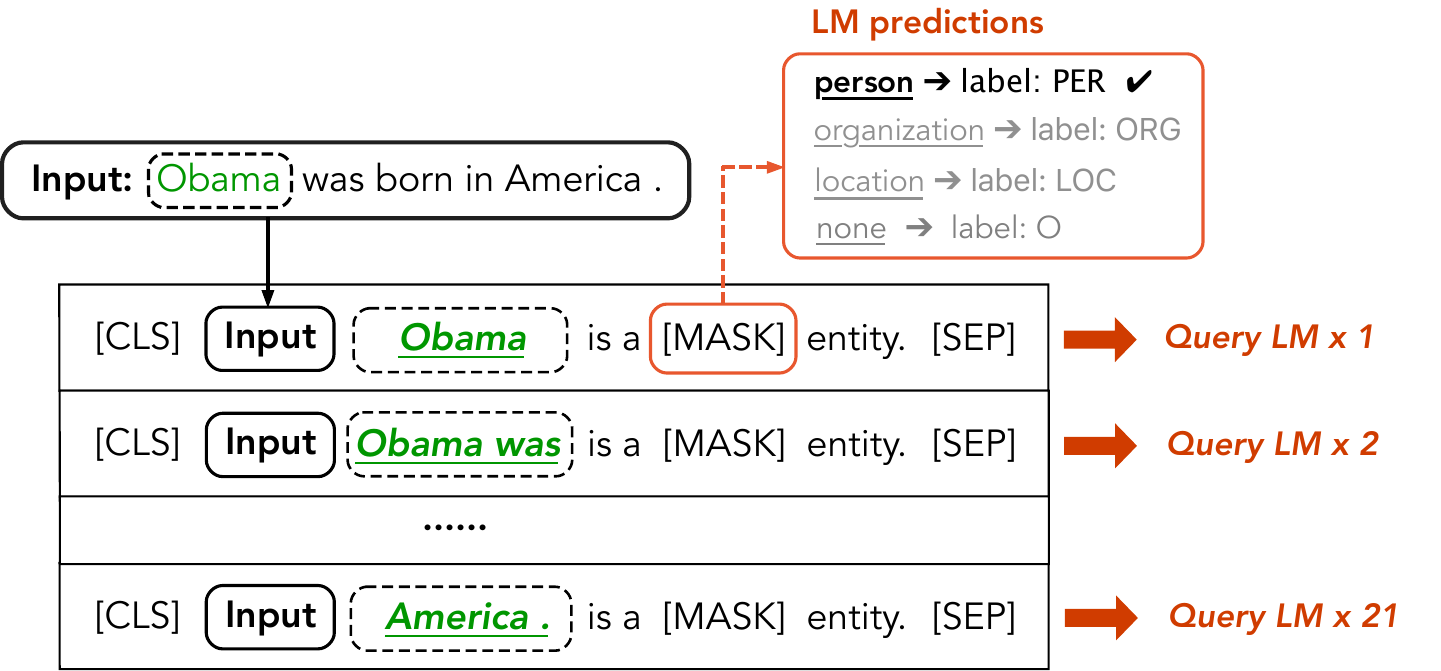}
    \caption{An example of template-based prompt method for NER. Predicting all labels in sentence ``Obama was born in America." requires enumeration over all spans.}
    \label{fig:template}
    \vspace{-0.3cm}
\end{figure}

Two main factors contribute to the success of prompt-based learning on few-shot classification. 
First, re-using the masked LM objective helps alleviate the gap between different training objectives used at pre-training and fine-tuning. Therefore, the LMs can faster adapt to downstream tasks even with a few training samples \cite{schick-schutze-2021-exploiting, schick-schutze-2021-just, NEURIPS2020_1457c0d6}.
Second, the sophisticated template and label word design helps LMs better fit the task-specific answer distributions, which also benefits few-shot performance. As proved in previous works, proper templates designed by manually selecting \cite{schick-schutze-2021-exploiting, schick-schutze-2021-just}, gradient-based discrete searching \cite{shin-etal-2020-autoprompt}, LM generating \cite{gao-etal-2021-making} and continuously optimizing \cite{DBLP:journals/corr/abs-2103-10385} are able to induce the LMs to predict more appropriate answers needed in corresponding tasks.

However, the template-based prompt methods are intrinsically designed for sentence-level tasks, and they are difficult to adapt to token-level classification tasks such as named entity recognition (NER). 
First, searching for appropriate templates is harder as the search space grows larger when encountering span-level querying in NER. What's worse, such searching with only few annotated samples as guidance can easily lead to overfitting.
Second, obtaining the label of each token requires enumerating all possible spans, which would be time-consuming. 
As an example in Fig.\ref{fig:template}, the input ``Obama was born in America." can be converted into ``Obama was born in America. [Z] is a \_\_ entity.", where [Z] is filled by enumerating all the spans in [X] (e.g., ``Obama", ``Obama was") for querying. Fig.\ref{fig:template} shows that obtaining all entities in ``Obama was born in America ." requires totally 21 times to query the LMs with every span.
Moreover, the decoding time of such an approach would grow catastrophically as sentence length increasing, making it impractical to document-level corpus.

In this work, we propose a more elegant way for prompting NER without templates.
Specifically, we reformulate NER as an LM task with an Entity-oriented LM (EntLM) objective. Without modifying the output head, the pre-trained LMs are fine-tuned to predict class-related pivot words (or label words) instead of the original words at the entity positions, while still predicting the original word at none-entity positions. 
Next, similar to template-based methods, we explore principled ways to automatically search for the most appropriate label words. Different approaches are investigated including selecting discrete label words based on the word distribution in lexicon-annotated corpus or LM predictions, and obtaining the prototypes as virtual label words.
Our approach keeps the merits of prompt-based learning as no new parameters are introduced during fine-tuning. Also, through the EntLM objective, the LM are allowed to perform NER task with only a slight adjustment of the output distribution, thus benefiting few-shot learning.
Moreover,
well-selected label words accelerate the adaptation of LM distribution towards the desired predictions, which also promotes few-shot performance. 
It's also worth noting that the proposed method requires only one-pass decoding to obtain all entity labels in the sentence, which is significantly more efficient compared to the time-consuming enumeration process of template-based methods. Our codes are publicly available at \url{https://github.com/rtmaww/EntLM/}.

To summarize the contribution of this work:
\begin{itemize}
\setlength{\itemindent}{0em}
\setlength{\itemsep}{0em}
\setlength{\topsep}{-0.6em}
    \item We propose a template-free approach to prompt NER under few-shot setting.
    \item We explore several approaches for label word engineering accompanied with intensive experiments.
    \item Experimental results verify the effectiveness of the proposed method under few-shot setting. Meanwhile, the decoding speed of the proposed method is 1930.12 times faster than template-based baseline.
\end{itemize}

\begin{figure*}
    \includegraphics[width=1.0\linewidth]{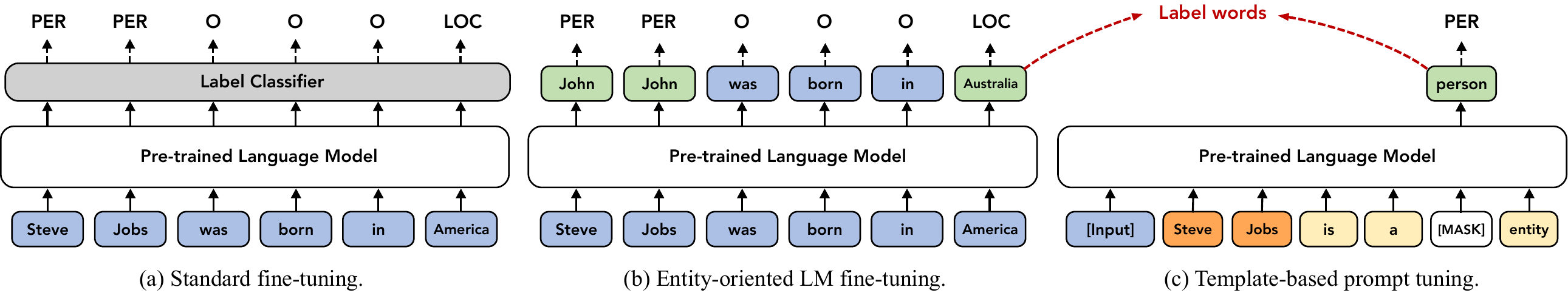}
    \caption{Comparison of different fine-tuning methods for NER. (a) is the standard fine-tuning method, which replace the LM head with a classifier head and perform label classification. (c) is the template-based prompt learning method, which induces the LM to predict label words by constructing a template. (b) is the proposed Entity-oriented LM fine-tuning method, which also re-uses the LM head and leads the LM to predict label words through an Entity-oriented LM objective. (For entities with multiple spans, the model predicts the same label word at each position, which is similar to the ``IO" labeling scheme.) }
    \label{fig:main_figure}
    \vspace{-0.3cm}
\end{figure*}

\section{Problem Setup}\label{task_description}

In this work, we focus on few-shot NER task.
Different from previous works that assume a rich-resource source domain and available support sets during testing, we follow the few-shot setting of \cite{gao-etal-2021-making}, which supposes that only a small number of examples are used for fine-tuning. Such setting makes minimal assumptions about available resources and is more practical. Specifically, when training on a new dataset $\mathbf{D}$ with the label space $\mathcal{Y}$,  we assume only $\textit{K}$ training examples for each class in the training set, such that the total number of examples is $K_{tot}=K\times |\mathcal{Y}|$. Then, the model is tested with an unseen test set $({X}^{test}, {Y}^{test})\sim \mathbf{D}_{test}$. Here, for NER task, a training sample refers to a continual entity span $\textbf{e}=\{x_1,\dots,x_m\}$ that is labeled with a positive class (e.g.,``PERSON").

\section{Approach}
In this work, we propose a template-free prompt tuning method, Entity-oriented LM (EntLM) fine-tuning, for few-shot NER. We first give a description of the template-based prompt tuning. Then we introduce the EntLM method along with the label word engineering process.

\subsection{Template-based Prompt Tuning}
The standard fine-tuning process for NER is replacing the LM head with a token-level classification head and optimizing the newly-introduced parameters and the pre-trained LM.
Different from standard fine-tuning, prompt-based tuning reformulates classification tasks as LM tasks, and fine-tunes LM to predict a label word. 

Formally, a prompt consists of a template function $T_{prompt}(\cdot)$ that converts the input $x$ to a prompt input $x_{prompt}=T_{prompt}(x)$, and a set of label words $\mathcal{V}$ which are connected with the label space through a mapping function $\mathcal{M} : \mathcal{Y} \to \mathcal{V}$. The template is a textual string with two unfilled slot: a input slot [X] to fill the input $x$ and an answer slot [Z] that allows LM to fill label words. For instance, for a sentiment classification task, the template can take the form as ``[X] It was [Z].". The input is then mapped to ``$x$ It was [Z].". Specifically, when using a masked language model (MLM) for prompt-based tuning, [Z] is filled with a mask token [MASK]. By feeding the prompt into the MLM, the probability distribution over the label set $\mathcal{Y}$ is modeled by:

\begin{equation}
\setlength\abovedisplayskip{0pt}
\setlength\belowdisplayskip{0pt}
\begin{aligned}
    P(y|x) &= P([MASK]=\mathcal{M}(\mathcal{Y})|x_{prompt}) \\
    &=Softmax(\mathbf{W}_{lm} \cdot \mathbf{h}_{[MASK]})
\end{aligned}
\end{equation}
where $\mathbf{W}_{lm}$ are the parameters of the pre-trained LM head. Unlike in standard fine-tuning, no new parameters are introduced in this approach, therefore the model can easier fit the target task with few samples. Also, the LM objective reduce the gap between pre-training and fine-tuning, thus benefiting few-shot training \cite{gao-etal-2021-making}.

\subsubsection{Problems of Prompt-based NER} 
However, when applied to NER, such prompt-based approach becomes complicated. given an input $X=\{x_1, \dots, x_n\}$, we need to obtain the label sequence $Y=\{y_1, \dots, y_n\}, y_i \in \mathcal{Y}$ corresponding to each token of $X$. Therefore, an additional slot [S] is added in the template to fill a token $x_i$ or a continual span $\mathbf{s}^i_j=\{x_i, \dots, x_j\}$ that starts from $x_i$ and ends with $x_j$. For example, the template can take the form as ``[X] [S] is a [Z] entity.", where the LMs are fine-tuned to predict an entity label word at [Z] (e.g., person) corresponding to an entity label (e.g., PERSON).
During decoding, obtaining the labels $Y$ of the whole sentence requires enumeration over all the spans:
\begin{equation}
\setlength\abovedisplayskip{0pt}
\setlength\belowdisplayskip{0pt}
\begin{split} 
    Y=\{\arg \max_{y\in\mathcal{Y}} P([Z]=\mathcal{M}(\mathcal{Y})|T_{prompt}(X,s^i_{j})), \\ s^i_j=Enumerate(\{x_i,\dots,x_j\}, i,j\in \{1..n\})\} ,
    \end{split}
\end{equation}
Such a decoding way is time-consuming and the decoding time increasing as the sequence length getting longer. Therefore, although efficient in few-shot setting, template-based prompt tuning is not suitable for NER task.

\subsection{Entity-Oriented LM Fine-tuning}
In this work, we propose a more elegant way to prompt NER without templates, while maintaining the advantages of prompt-tuning.
Specifically, we also reformulate NER as a LM task. However, instead of forming templates to re-use the LM objective, we propose a new objective, Entity-oriented LM (EntLM) objective for fine-tuning NER. As shown in Fig. \ref{fig:main_figure} (b), when fed with ``Obama was born in America", the LM is trained to predict a label word ``John" at the position of the entity ``Obama" as an indication of the label ``PER". While for none-entity word ``was", the LM remains to predict the original word.

Formally, to fine-tune the LM with EntLM objective,
we first construct a label word set $\mathcal{V}_l$ which is also connected with the task label set through a mapping function $\mathcal{M} : \mathcal{Y} \to \mathcal{V}_l$. 
Next, given the input sentence $X=\{x_1, \dots, x_n\}$ and the corresponding label sequence $Y=\{y_1, \dots, y_n\}$, we construct a target sentence $X^{Ent}=\{x_1, \dots, \mathcal{M}(y_i), \dots, x_n\}$ by replacing the token at the entity position $i$ (here we assume $y_i$ is an entity label) with corresponding label word $\mathcal{M}(y_i)$, and maintaining the original words at none-entity positions. Then, given the original input $X$, the LM is trained to 
maximize the probability $P(X^{Ent}|X)$ of the target sentence $X^{Ent}$:

\begin{equation}
\setlength\abovedisplayskip{0pt}
\setlength\belowdisplayskip{2pt}
        \mathcal{L}_{EntLM}=-\sum_{i=1}^n log P(x_i=x_i^{Ent}|X)
\end{equation}
where $P(x_i=x_i^{Ent}|X)= Softmax(\mathbf{W}_{lm} \cdot \mathbf{h}_{i})$. Noted that $\mathbf{W}_{lm}$ are also the parameters of the pre-trained LM head. By re-using the whole pre-trained model, no new parameters are introduced during this fine-tuning process. Meanwhile, the EntLM objective serves as a LM-based objective to reduce the gap between pre-training and fine-tuning. In this way, we avoid the complicated template constructing for NER task, and keep the good few-shot ability of prompt-based method.

During testing, we directly feed the test input $X$ into the model, and the probability of labeling the $i^{th}$ token with class $y \in \mathcal{Y}$ is modeled by:

\begin{equation}
\begin{aligned}
    p(y_i=y|X) &= p(x_i=\mathcal{M}(y)|X) \\
\end{aligned}
\end{equation}
Noted that we only need one-pass decoding process to obtain all labels for each sentence, which is intensively more efficient than template-based prompt querying.

\subsection{Label Word Engineering}\label{selectlabelword}

Previous template-based studies have verified the significant impact of template engineering on few-shot performance. Similarly, in this work, we explore approaches for automatically selecting proper label words. Since the EntLM object lead all entities that belong to a class to predict the same label word, we believe that the purpose of label word searching is to find a pivot word that can mostly represent the words in each class. 

\subsubsection{Low-resource Label word selection }
When selecting label words with only few annotated samples as guidance, the randomness of sampling will largely affect the selection. In order to obtain more consistent selection, we explore the usage of unlabeled data and lexicon-based annotation as a resource for label word searching. This is a practical setting since unlabeled data of a target domain or a general domain is usually available, and for NER, the entity lexicon of target classes are usually easy to access.

\begin{figure}
    \includegraphics[width=1.0\linewidth]{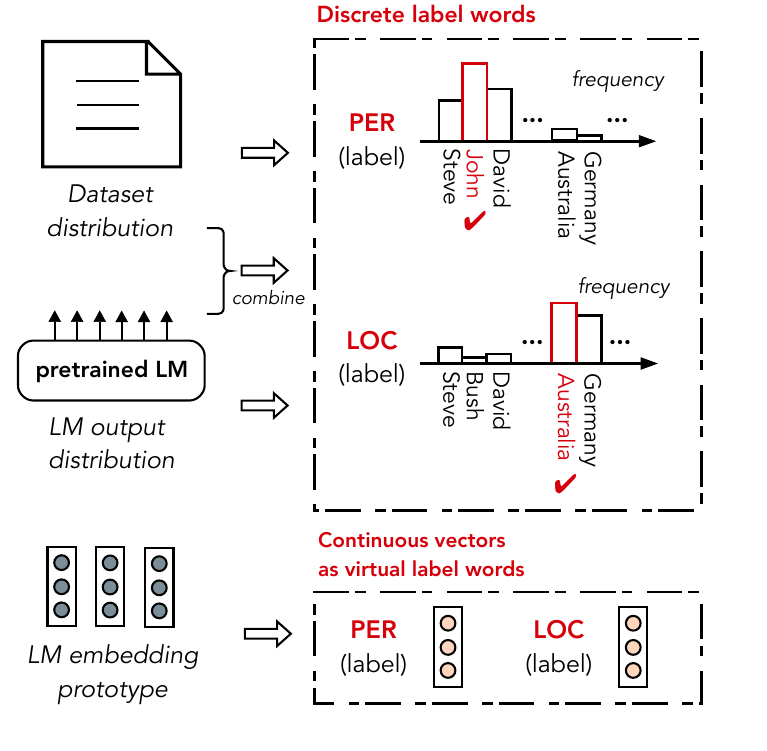}
    \caption{Searching for two types of label words: the discrete label words and the continuous vectors as virtual label words. To search for the discrete label words, we select the high-frequency words in data or LM output distribution, or combine these two ways. To search for virtual label words, we calculate the mean vectors of the high-frequency words of each class as prototypes. }
    \label{fig:baglabel}
    \vspace{-0.3cm}
\end{figure}

To obtain annotation via entity lexicon, we adopt the KB-matching approach proposed by \citet{10.1145/3394486.3403149}, which leverages an external KBs, wikidata, as the source of lexicon annotation. Such lexicon-based annotation is inevitably noisy. However, our approach do not suffers a lot from the noise since we only regarded it as an indication of the data distribution and do not train the model directly with the noisy annotation.

\subsubsection{Label word searching}
With the help of lexicon-annotated data $\mathcal{D}_{lexicon}=\{(X_i, Y_i^*)\}^N_{i=1}$, we explore three methods for label word searching. 

\noindent \textbf{Searching with data distribution (Data search)}  The most intuitive method is to select the most frequent word of the given class in the corpus. Specifically, when searching for label words for class $C$, we calculate the frequency $\phi(x=w,y^*=C)$ of each word $w \in \mathcal{V}$ labeled as $C$ and select the most frequent words by ranking:
    \begin{equation}\label{data_D}
        \mathcal{M}(C)=\arg\max_w \phi(x=w,y^*=C)
    \end{equation}

\noindent \textbf{Searching with LM output distribution (LM search)} In this approach, we leverage the pre-trained language model for label word searching. Specifically, we feed each sample $(X, Y^*)$ into LM and get the probability distribution $p(\hat{x}_i=w|X)$ of predicting each word $w \in \mathcal{V}$ at each position $j$.
    Suppose $\mathcal{I}_{topk}(\hat{x}_i=w|X,Y^*) \to \{0,1\}$ is the indicator function indicating whether $w$ belongs to the top$k$ predictions of $x_i$ in sample $(X,Y^*)$. The label word of class $C$ can be obtained by: 
    \begin{equation}\label{LMoutput_D}
    \setlength\abovedisplayskip{0pt}
\setlength\belowdisplayskip{0pt}
    \begin{split}
        \mathcal{M}(C)=\arg\max_w & \\\sum_{\scriptscriptstyle (X,Y^*)\in \mathcal{D}} & \!\sum_i^{|X|} \phi_{topk}(\hat{x}_i=w,y_i^*=C)
        \end{split} 
    \end{equation}
    where $\phi_{topk}(\hat{x}_i=w,y^*=C)=\mathcal{I}_{topk}(\hat{x}_i=w|X,Y^*)\cdot\mathcal{I}(y_i^*=C)$ denotes the frequency of $w$ occurring in the top $k$ predictions of the positions labeled as class $C$. 

\noindent \textbf{Searching with both data \& LM output distribution (Data\&LM seach)} In this approach, we select label words by simultaneously considering the data distribution and LM output distribution. Specifically, the label word of class $C$ can be obtained by: 
    \begin{equation}\label{both_D}
    \setlength\abovedisplayskip{-1pt}
\setlength\belowdisplayskip{-3pt}
    \begin{split}
        \mathcal{M}(C)=\arg\max_w \{ \sum_{\scriptscriptstyle(X,Y^*)\in \mathcal{D}}\! \sum_i^{|X|} \phi(x_i=w,y_i^*=C) \\
        \cdot \sum_{\scriptscriptstyle(X,Y^*)\in \mathcal{D}}\! \sum_i^{|X|}
        \phi_{topk}(\hat{x}_i=w,y_i^*=C)\}
        \end{split} 
    \end{equation}

\noindent \textbf{Virtual label word (Virtual)} Instead of using real words, in this approach, we search for continuous vectors that can be regarded as virtual label words. One intuitive way is to follow the practice of Prototypical Networks \cite{NIPS2017_cb8da676}, which uses the mean vector of the embeddings of words belonging to each class as a prototype. Since averaging the embeddings of all the words belong to a class is expensive, here we simply use the mean vector of the top$k$ high-frequency words selected by the previous approaches:
        \begin{equation}
        \mathcal{M}(C)=\frac{1}{|\mathcal{V}_C|}\sum_{w\in\mathcal{V}_C}f_{\phi}(w)
    \end{equation}
    where $\mathcal{V}_C$ is the set of label words obtaining by finding the top $k$ words with Eq. \ref{data_D},\ref{LMoutput_D},\ref{both_D}, and $f_{\phi}(\cdot)$ denotes the embedding function of the pre-trained model.

\subsubsection{Removing conflict label words}
The selected high-frequency label words are potentially high-frequency words among all the classes. Using such label words will result in conflicts when training for different classes. Therefore, after label word selection, we remove the conflict label words of a class $C$ by:
\begin{equation}
    w=\mathcal{M}(C), if    \frac{\phi(x=w,y^*=C)}{\sum_k \phi(x=w,y^*=k)} > Th
\end{equation}
where $Th$ is a manually set threshold.

\begin{table}[]
\centering
\small

\begin{tabular}{lccccc}
\toprule
\textbf{Datasets} & \textbf{Domain} & \textbf{\# Class} & \textbf{\# Train} & \textbf{\# Test} \\
\midrule
CoNLL'03 & News & 4 & 14.0k & 3.5k \\
OntoNotes*  & General & 11 & 60.0k & 8.3k\\
MIT Movie & Review & 12 & 7.8k & 2.0k\\

\bottomrule

\end{tabular}
\caption{Dataset details. OntoNotes* denotes the Ontonotes5.0 dataset after removing value/numerical/time/date entity types.}
\label{tab:dataset}
\vspace{-0.3cm}
\end{table}

\begin{table*}[ht]
\centering
\small
\begin{tabular}{c|l|cccc}
\hline
\hline
\multicolumn{1}{c|}{\multirow{2}{*}{\textbf{Datasets}}} &
\multicolumn{1}{c|}{\multirow{2}{*}{\textbf{Methods}}} &
\multicolumn{1}{c}{\multirow{2}{*}{\textbf{K=5}}} & \multicolumn{1}{c}{\multirow{2}{*}{\textbf{K=10}}} & \multicolumn{1}{c}{\multirow{2}{*}{\textbf{K=20}}} & \multicolumn{1}{c}{\multirow{2}{*}{\textbf{K=50}}} \\ 
\multicolumn{1}{c|}{} & \multicolumn{1}{c|}{} & \multicolumn{1}{c}{ }& 
\multicolumn{1}{c}{ } &
\multicolumn{1}{c}{ } & 
\multicolumn{1}{c}{}  \\ \cline{1-6}
\hline
\multirow{6}{*}{\textbf{CoNLL03}}&

BERT-tagger (IO) &  41.87 (12.12) & 59.91 (10.65) & 68.66 (5.13) & 73.20 (3.09) \\ %
&NNShot & 42.31 (8.92)& 59.24 (11.71) & 66.89 (6.09) & 72.63 (3.42) \\
&StructShot & 45.82 (10.30) & 62.37 (10.96) & 69.51 (6.46) & 74.73 (3.06)  \\

&Template NER& 43.04 (6.15) & 57.86 (5.68) & 66.38 (6.09) & 72.71 (2.13)  \\

&\cellcolor{mygray}EntLM (Ours)& \cellcolor{mygray} 49.59 (8.30) &\cellcolor{mygray}64.79 (3.86) &\cellcolor{mygray} 69.52 (4.48) &\cellcolor{mygray} 73.66 (2.06)   \\
&\cellcolor{mygray}EntLM + Struct (Ours)& \cellcolor{mygray}\textbf{51.32} (7.67) & \cellcolor{mygray}\textbf{66.86} (3.01) & \cellcolor{mygray}\textbf{71.23} (3.91) & \cellcolor{mygray}\textbf{74.80} (1.87)  \\
\hline
\hline

\multirow{6}{*}{\textbf{OntoNotes 5.0}}&

BERT-tagger (IO) &  34.77 (7.16) & 54.47 (8.31) & 60.21 (3.89) & 68.37 (1.72) \\ 

&NNShot & 34.52 (7.85)& 55.57 (9.20) & 59.59 (4.20) & 68.27 (1.54) \\
&StructShot & 36.46 (8.54) & 57.15 (5.84) & 62.22 (5.10) & 68.31 (5.72)  \\

&Template NER& 40.52 (8.62) & 49.89 (3.66) & 59.53 (2.25) & 65.15 (2.95)  \\

&\cellcolor{mygray}EntLM (Ours)& \cellcolor{mygray} {45.21} (9.17) &\cellcolor{mygray}57.64 (4.18) &\cellcolor{mygray} 65.64 (4.24) &\cellcolor{mygray} 71.77 (1.31)   \\
&\cellcolor{mygray}EntLM + Struct (Ours)&\cellcolor{mygray} \textbf{46.60} (10.35) & \cellcolor{mygray}\textbf{59.35} (3.24) &\cellcolor{mygray} \textbf{67.91} (4.55) & \cellcolor{mygray}\textbf{73.52} (0.97)  \\
\hline
\hline

\multirow{6}{*}{\textbf{MIT-Movie}}&

BERT-tagger (IO) &  39.57 (6.38) & 50.60 (7.29) & 59.34 (3.66) & 71.33 (3.04) \\ 

&NNShot & 38.97 (5.54)& 50.47 (6.09) & 58.94 (3.47) & 71.17 (2.85) \\
&StructShot & 41.60 (8.97) & 53.19 (5.52) & 61.42 (2.98) & 72.07 (6.41)  \\

&Template NER& 45.97 (3.86)& 49.30 (3.35) & 59.09 (0.35) & 65.13 (0.17)  \\

&\cellcolor{mygray}EntLM (Ours)&  \cellcolor{mygray}46.62 (9.46) &\cellcolor{mygray}57.31 (3.72) &\cellcolor{mygray} 62.36 (4.14)& \cellcolor{mygray}71.93 (1.68)   \\
&\cellcolor{mygray}EntLM + Struct (Ours)&\cellcolor{mygray} \textbf{49.15} (8.91) & \cellcolor{mygray}\textbf{59.21} (3.96) & \cellcolor{mygray} \textbf{63.85} (3.7) &\cellcolor{mygray} \textbf{72.99} (1.80)  \\
\hline
\hline

\end{tabular}
\caption{Main results of EntLM on three datasets under different few-shot settings (K=5,10,20,50). We report mean (and deviation in brackets) performance over 3 different splits (4 repeated experiments for each split).}
\label{tab:main_result}
\vspace{-0.3cm}
\end{table*}

\section{Experiments}
In this section, we conduct few-shot experiments to verify the effectiveness of the proposed method. We also conducts intensive analytical experiments for label words selection.

\subsection{Experimental settings}
As mentioned in Section \ref{task_description}, in this work, we focus on few-shot setting that no source domain data yet only $K$ samples of each class are available for training on a new NER task. To better evaluate the models' few-shot ability, we conduct experiments with $K\in \{5,10,20,50\}$. For each $K$-shot experiment, we sample 3 different training set and repeat experiments on each training set for 4 times.

\noindent \textbf{Few-shot data sampling.} Different from sentence-level few-shot tasks, in NER, a sample refers to one entity span in a sentence. One sampled sentence might include multiple entity instances.
In our experiments, we conduct an exact sampling strategy to ensure that we sample exactly $K$ samples for each class. The details of the algorithm can be found at Appendix \ref{sample_algo}.

\subsection{Datasets and Implementation Details}
We evaluate the proposed method with three benchmark NER datasets from different domains: the CoNLL2003 dataset \cite{sang2003introduction} from the newswire domain, Ontonotes 5.0 dataset \cite{weischedel2013ontonotes} from general domain and the MIT-Movie dataset \cite{liu2013query}\footnote{https://groups.csail.mit.edu/sls/downloads/} from the review domain. As we focus on named entities, we omit the value/numerical/time/date entity types (e.g.,``Cardinal", ``Money", etc) in OntoNotes 5.0. Details of the datasets are shown in Table \ref{tab:dataset}.

 \textbf{Labeling multi-span entities.} For entities with multiple spans (including multiple words or sub-tokens after tokenization), we let the model predict the same label word at each position. This labeling method is the same with the ``IO" labeling schema, which is consistent to our baseline implementation.

To ensure a few-shot scenario, we didn't use a development set for model choosing. Instead, we use the model of the last epoch for predicting. For lexicon-based annotation, we use the KB-matching method of \citet{10.1145/3394486.3403149}\footnote{https://github.com/cliang1453/BOND}. For more implementation details (e.g., the learning rate, etc.), please refer to Appendix \ref{details} or our codes.

\subsection{Baselines and Proposed Models}
In our experiments, we compare our method with competitive baselines, involving both metric-learning based and prompt-based approaches.

\textbf{BERT-tagger} \cite{devlin-etal-2019-bert} The BERT-based baseline which fine-tunes the BERT model with a label classifier.

\textbf{NNShot} and \textbf{StructShot} \cite{yang-katiyar-2020-simple} Two metric-based few-shot learning approaches for NER. Different from Prototypical Network, they leverage a a nearest neighbor classifier for few-shot prediction. StructShot is an extension of NNShot which proposes a viterbi algorithm during decoding. We extend these two approaches to our few-shot setting. Noted that the viterbi algorithm in the original paper calculates the data distribution of a source domain, yet in our setting, the source domain is unavailable. Therefore, we also use the \textbf{lexicon-annotated} data for performing this method.

\textbf{TemplateNER} \cite{cui-etal-2021-template} A template-based prompt method. By constructing a template for each class, it queries each span with each class separately. The score of each query is obtained by calculating the generalization probability of the query sentence through a generative pre-trained LM, BART\cite{lewis-etal-2020-bart}.

\textbf{EntLM} The proposed method.

\textbf{EntLM+Struct} Based on the proposed method, we further leverages the viterbi algorithm proposed in \cite{yang-katiyar-2020-simple} to boost the performance. For more details please refer to \cite{yang-katiyar-2020-simple} or our codes.

In Appendix \ref{comprehensive}, we also compare with the roberta-base baselines from \cite{huang2020fewshot}.

\subsection{Few-shot Results}

Table \ref{tab:main_result} show the results of the proposed method and baselines under few-shot setting. From the table, we can observe that: (1) On all the three datasets, for all few-shot settings, the proposed method performs consistently better than all the baseline methods, especially for 5-shot learning. Also, the performance of the proposed method is more stable (according to the deviation) than the compared baselines. (2)  BERT-tagger method shows poor ability of few-shot learning, and the proposed method achieves up to 9.45\%, 11.83\%, 9.58\% improvement over BERT-tagger on CoNLL03, OntoNotes 5.0 and MIT-Movie, respectively. These results show the advantages of the proposed method over standard fine-tuning, which introduces no new parameters and uses an LM-like objective to reduce the gap between pre-training and fine-tuning. 
(3) The proposed method consistently outperforms the template-based prompt method, Template NER, which shows the advantage of the proposed method over standard template-based method. 
(4) When no rich-resource source domain is available, the metric-based methods (NNShot) do not show advantages over BERT-tagger, which shows the limitation of these method under more practical few-shot scenarios. (5) Among all baselines, the StructShot is a competitive baseline that also leverages lexicon and unlabeled data for structure-based decoder, yet our method can also benefit from the viterbi decoder and outperform StructShot.

\begin{table*}[]
\centering
\small

\begin{tabular}{lcccccc}
\toprule
\multirow{2}{*}{\textbf{Methods}} & \multicolumn{2}{c}{\textbf{CoNLL03}} & \multicolumn{2}{c}{\textbf{OntoNotes}} & \multicolumn{2}{c}{\textbf{MIT-Movie}} \\
\cline{2-7}
&{K=5} &  {K=10} & {K=5} &
{K=10} &{K=5} &
{K=10} \\
\midrule

DataSearch & \textbf{50.00} (9.75) &  61.31 (4.73) & 36.94 (5.04) & 49.54 (5.02)& 39.25 (4.83) & 51.65 (5.52)\\ 
LMSearch  & 48.40 (6.81) &  59.39 (5.50) & 36.98 (6.71) & 48.20 (5.46)&39.12 (4.18)& 48.30 (3.76)\\
Data\&LMSeach & 49.55 (7.76) & 61.00 (6.98)& 36.60 (7.90) & 50.64 (6.12)& 38.86 (11.43)& 50.42 (6.45) \\                     
Data + Virtual & 49.25 (4.96) & 63.40 (5.13) & \textbf{45.61} (10.51) & 55.13 (4.95) & 45.59 (8.25)& 55.10 (4.42)\\
LM + Virtual & 42.65 (12.58) & 59.39 (5.50)& {45.29} (7.77) & 54.50 (3.66) & 46.23 (5.60)& 54.92 (6.15)\\
Data\&LM + Virtual & {49.59} (8.30) & \textbf{64.79} (3.86) & {45.21} (9.17) &\textbf{57.64} (4.18) & \textbf{46.62} (9.46)& \textbf{57.31} (3.72)\\

\bottomrule

\end{tabular}
\caption{Comparison of our label word selection methods. We report mean (and standard deviation) performance.}
\label{tab:labelwordexpr}
\vspace{-0.2cm}
\end{table*}

\begin{table}[]
\centering
\small

\begin{tabular}{lccc}
\toprule
\textbf{Methods} & \textbf{CoNLL} & \textbf{OntoNotes} & \textbf{MIT-Movie}  \\
\midrule
BERT-tagger & 8.57 & 23.89 & 6.46  \\
TemplateNER  & 6,491.00 & 50,241.00 & 5254.00 \\
NNShot & 16.03 & 82.62 & 15.98 \\                     
StructShot & 19.84 & 98.67 & 17.66 \\
EntLM & 9.26 & 26.03 & 6.64\\
EntLM + Struct & 13.40 & 34.92& 7.38 \\

\bottomrule

\end{tabular}
\caption{The decoding time (s) of different methods.}
\label{tab:speed}
\vspace{-0.3cm}
\end{table}

\subsection{Efficiency Study}
In this section, we perform an efficiency study on all the three datasets. We calculate the decoding time of each method on a TiTan XP GPU with batch size=8. (The source codes of Template NER do not allow us to change the batch size, so we keep the original batch size=45, which is the enumeration number of a 9-gram span. ) From Tab.\ref{tab:speed}, we can observe that: 1) EntLM can achieve comparable speed with BERT-tagger, as only one pass of token classification is required for decoding each batch. 2) The decoding speed of TemplateNER is severely slow, while EntLM is up to 1930.12 times faster than TemplateNER. These results show the advantages of EntLM over template-based prompt tuning methods in NER task.

\begin{figure}
    \centering
    \includegraphics[width=0.8\linewidth]{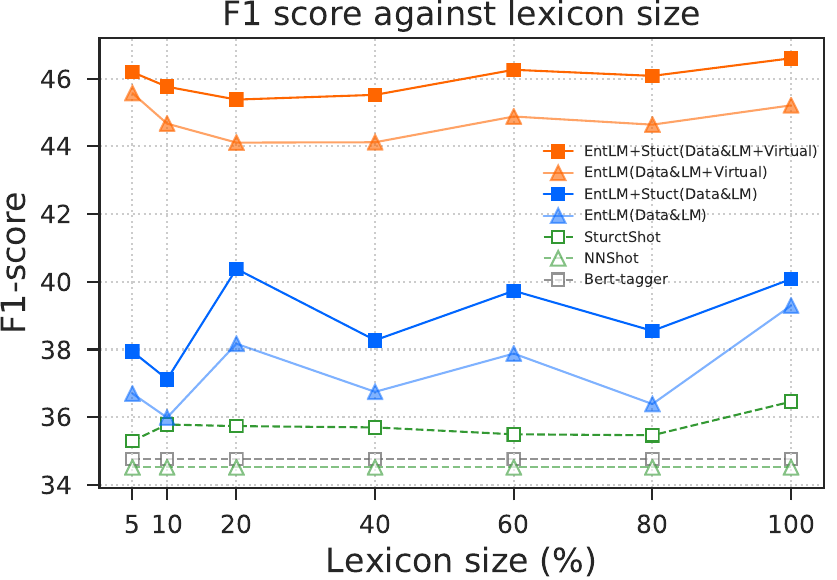}
    \caption{Impact of different lexicon sizes.}
    \label{fig:lexicon_control}
    \vspace{-0.2cm}
\end{figure}

\subsection{Label Word Selection}

In Sec.\ref{selectlabelword}, we have presented different ways for label word selection. In this section, we conduct experiments on these methods and the results are reported in table \ref{tab:labelwordexpr}. We can observe that: 1) The virtual word selection approach is always better than the discrete word selection. While among all virtual selection methods, choosing high-frequency words with the combination of data and LM distribution shows advantages over other methods. The reason of these results might be that simultaneously considering both data distribution gives not only the data prior in the target dataset, but also the contextualized information from the PLM, thus benefiting the performance. 2) Searching only with LM distribution leads to poor results especially under 5-shot setting, showing that the general knowledge learned from pre-trained might be less helpful than the data-specific knowledge under few-shot settings.

\subsubsection{Impact of Lexicon Quality on Label Word Selection}
Note that we leverage unlabeled data and lexicon annotation for label word selection. In this experiment, we study how the quality of lexicon impacts the performance on the OntoNotes* dataset. Specifically, we obtain different sizes of lexicon (5\% to 80\% of the original lexicon size) by sampling entity words in the original lexicon with the weights of entity frequency. This sampling method follows the real-world situation since high-frequency entities are easier to obtain. Fig.\ref{fig:lexicon_control} shows the results of EntLM and baseline methods against lexicon size. We can observe that: (1) EntLM with the Data\&LM+Virtual selection method illustrates consistent high performance even with 5\% lexicon. This means our method is not limited to the lexicon quality, and we only require a small lexicon to reach acceptable few-shot performance. (2) Compared with Data\&LM+Virtual method, the Data\&LM is much more fragile regarding the lexicon quality. However, it still performs better than the compared baselines.

\begin{figure}
    \centering
    \includegraphics[width=0.8\linewidth]{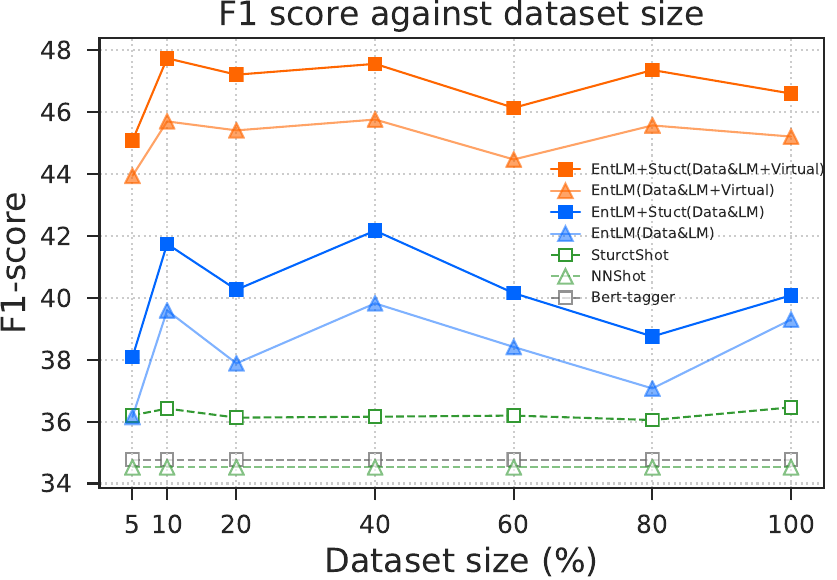}
    \caption{Impact of the amount of unlabeled data.}
    \label{fig:data_control}
    \vspace{-0.1cm}
\end{figure}

We further conduct experiments on different sizes of the unlabeled dataset by uniformly sampling 5\%-80\% of the original data. As shown in Fig.\ref{fig:data_control}, the proposed method also shows high robustness to the amount of unlabeled data.

\renewcommand\arraystretch{0.9}
\begin{table}[]
\centering
\small

\begin{tabular}{lcccc}
\toprule
\multirow{2}{*}{\textbf{Methods}} & \multicolumn{2}{c}{\textbf{CoNLL03}}  \\
\cline{2-3}
&{K=5} &  {K=10} \\
\midrule

BERT-tagger &41.87 (12.12) & 59.91 (10.65)    \\ 

EntLM &49.59 (8.30) &64.79 (3.86) \\  
EntLM + Struct &{51.32} (7.67) & {66.86} (3.01)  \\ 
BERT-tagger (further)  & 41.16 (10.41) &  61.70 (5.15) \\
EntLM (further) & 56.82 (12.27) & 66.82 (4.65) \\     
EntLM + Struct (further) &58.77 (12.16) & 68.96 (4.41) \\

\bottomrule

\end{tabular}
\caption{Impact of further pre-training.}
\label{tab:further}
\vspace{-0.3cm}
\end{table}

\subsection{Effect of Further Pre-training}
When predicting label words on task-specific data during fine-tuning, there is an intrinsic gap between the LM output distribution and the target data distribution. Therefore, it is natural to conduct a further pre-training approach on the target-domain unlabeled data to boost the LM predictions towards target distribution. In Table \ref{tab:further}, we show the results of our method and BERT-tagger trained after further pre-training with MLM objective on domain-specific unlabeled data. As seen, the further pre-training practice can largely boost the few-shot learning ability of EntLM, while showing less helpful for classifier-based fine-tuning method. This might because the LM objective used in EntLM can benefit more from a task-specific LM output distribution, showing the superiority of EntLM in better leveraging the pre-trained models.

\section{Related Works}

\subsection{Template-based prompt learning}
Stem from the GPT models \cite{radford2019language,NEURIPS2020_1457c0d6},  prompt-based learning have been widely discussed.
These methods reformulate downstream tasks as cloze tasks with textual templates and a set of label words, and the design of templates is proved to be significant for prompt-based learning. \citet{schick-schutze-2021-exploiting,schick-schutze-2021-just} uses manually defined templates for prompting text classification tasks. \citet{jiang-etal-2020-know} proposes a mining approach for automatically search for templates. \citet{shin-etal-2020-autoprompt} searches for optimal discrete templates by a gradient-based approach. \cite{gao-etal-2021-making} generates templates with the T5 pre-trained model. Meanwhile, several approaches have explore continuous prompts for both text classification and generation tasks \citet{li-liang-2021-prefix, DBLP:journals/corr/abs-2103-10385, han2021ptr}. Also, several approaches are proposed to enhance the templates with illustrative cases \cite{madotto2020language, gao-etal-2021-making,NEURIPS2020_1457c0d6} or context \cite{petroni2020how}. Although template-based methods are proved to be useful in sentence-level tasks, for NER task \cite{cui-etal-2021-template}, such template-based method can be expensive for decoding. Therefore, in this work, we propose a new paradigm of prompt-tuning for NER without templates.

\subsection{Few-shot NER}

Recently, many studies focuses on few-shot NER \cite{hofer2018fewshot,Fritzler_2019,9262018,ding-etal-2021-nerd,chen2021lightner}. Among these, \citet{Fritzler_2019} leverages prototypical networks for few-shot NER. \citet{yang-katiyar-2020-simple} propose to calculate the nearest neighbor of each queried sample instead of the nearest prototype. \citet{huang-etal-2021-shot} experimented comprehensive baselines on different datasets. \citet{tong-etal-2021-learning} proposes to mine the undefined classes for few-shot learning. \citet{cui-etal-2021-template} leverages prompts for few-shot NER. However, most of these studies follow the manner of episode training or assume a rich-resource source domain. In this work, we follow the more practical few-shot setting of \citet{gao-etal-2021-making}, which assumes only few samples each class for training. We also adapt previous methods to this setting as competitive baselines.

\section{Conclusion}
In this work, we propose a template-free prompt tuning method, EntLM, for few-shot NER. Specifically, we reformulate the NER task as a Entity-oriented LM task, which induce the LM to predict label words at entity positions during fine-tuning. In this way, not only the complicated template-based methods can be discarded, but also the few-shot performance can be boosted since the EntLM objective reduces the gap between pre-training and fine-tuning. Experimental results show that the proposed method can achieve significant improvement on few-shot NER over BERT-tagger and template-based method. Also, the decoding speed of EntLM is up to 1930.12 times faster than the template-based method.

\section*{Acknowledgements}
The authors wish to thank the anonymous reviewers for their helpful comments. This work was partially funded by National Natural Science Foundation of China (No. 61976056, 62076069), Shanghai Municipal Science and Technology Major Project (No.2021SHZDZX0103).

\bibliography{anthology,custom}
\bibliographystyle{acl_natbib}

\newpage

\appendix

\section{Appendix}

\subsection{Implementation Details}\label{details}
We implement our method based on the huggingface transformers\footnote{https://github.com/huggingface/transformers}. For all our experiments except TemplateNER, we use ``bert-base-cased" pre-trained model as the base model for fine-tuning, and no new parameters are introduced in the proposed method. For both bert-base baselines and our method, we set learning rate=1e-4 and batch size=4 for few-shot training. For all experiments, we train the model for 20 epochs, and AdamW optimizer is used with the same linear decaying schedule as the pre-training stage. These hyper-parameter settings are as the same with \cite{huang-etal-2021-shot}. For other hyper-parameter settings of the baseline methods, we simply follow the default settings. 
 When implementing all methods, we adopt the ``IO" labeling schema since we found that the ``IO" schema is better than ``BIO" schema under few-shot setting.

As for label word selection, we use the Data\&LM seaching along with the virtual method (Data\&LM+Virtual) for all dataset and set the conflict ratio to $Th=0.6$. When selecting the top $k$ high-frequency words for virtual method, we set k to 6.

\subsection{Sampling Algorithm}\label{sample_algo}
We conduct an exact sampling algorithm to ensure sampling exactly $K$ samples for each class, which is different from the greedy sampling method used in previous methods \cite{yang-katiyar-2020-simple}. The algorithm is detailed in Algorithm \ref{algo:support}. For all of the three datasets we used, we exactly obtained $K$ samples for each class under all the $K$-shot setting.

\subsection{Effect of Conflict threshold}
\begin{figure}
    \centering
    \includegraphics[width=0.8\linewidth]{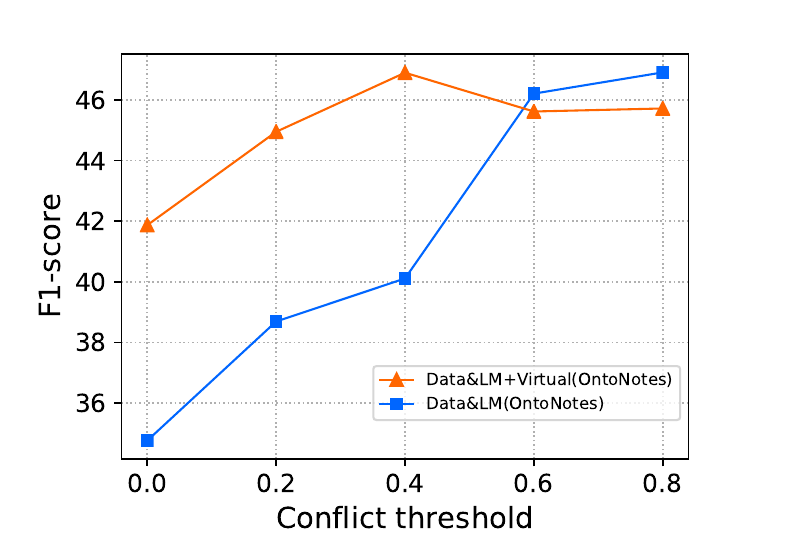}
    \caption{Impact of the conflict threshold. }
    \label{fig:conflict}
    \vspace{-0.2cm}
\end{figure}

Fig. \ref{fig:conflict} shows the impact of conflict threshold on 5-shot performance. As seen, for Data\&LMSearch, lower conflict threshold results in improper label words that bring noisy annotated entities. Therefore, the performance promoting as the conflict threshold increasing. As for Data\&LM+Virtual method, the impact of conflict words are less significant since multiple words are selected to construct the virtual vector.

\subsection{Effect of $k$ in virtual method}
\begin{figure}
    \centering
    \includegraphics[width=0.8\linewidth]{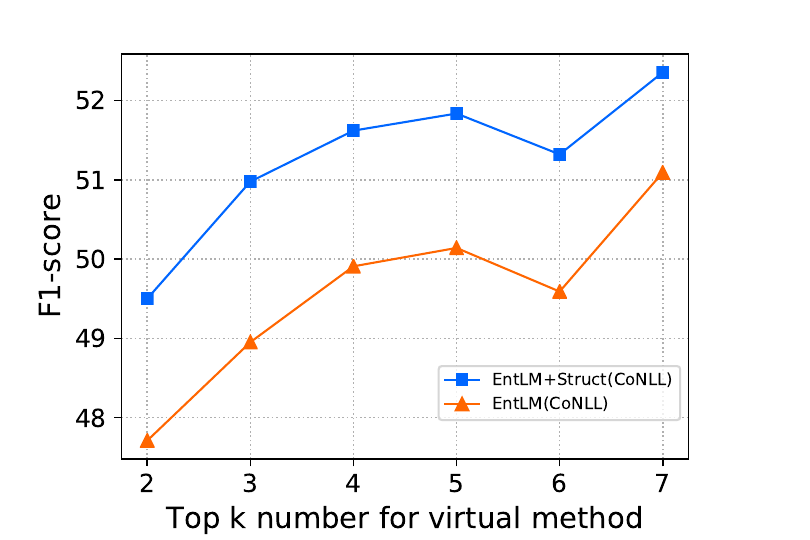}
    \caption{Effect of the choice of top $k$ number for virtual method. }
    \label{fig:topk}
    \vspace{-0.2cm}
\end{figure}

Fig.\ref{fig:topk} shows the impact of the choice of top $k$ number for virtual method. We conduct experiments using the Data\&LMSearch+Virtual method on CoNLL 5-shot dataset. We can see that the performance of the proposed method is robust to the choice of $k$, since it can consistently achieve good results when $k>=3$. In our main experiments, we simply choose $k=6$ for all datasets.

\subsection{Comparison with Comprehensive few-shot NER benchmark}\label{comprehensive}
We also conduct experiments on the few-shot benchmark provided by \cite{huang-etal-2021-shot}, in order to compare with the competitive baselines in the paper. These methods are implemented with the ``Roberta-base" pretrained model. Therefore, we also implement our method based on ``Roberta-base" for fair comparison. Since the sampled data of OntoNotes is not available, we only experimented on the CoNLL'03 and MIT-Movie datasets. The results are shown in Table \ref{tab:comprehensive}.

The results show that, our method outperforms over all baselines. Notice that the NSP method leverages the 6.8GB WiFiNE dataset for pre-training, and that the ST method performs self-training on the unlabeled data. However, our method still shows better results, which illustrates the effectiveness of the proposed objective over standard fine-tuning. Also, the proposed method can be further boosted with NSP and ST. We leave this for future works.

\subsection{Case Study}
In Table \ref{tab:label_word_example}, we show the label words selected with the Data\&LM+Virtual method as examples.

\begin{table}[]\label{comprehensive_table}
\small
\centering

\begin{tabular}{lccc}
\toprule
\textbf{Methods} & \textbf{CoNLL}   & \textbf{MIT-Movie}\\
& 5-shot & 5-shot \\
\midrule
LC & 53.5 & 51.3  \\
LC+NSP  &61.4 & 53.1 \\
Proto & 58.4 &  38.0 \\                     
Proto+NSP & 60.9 &  43.8 \\
LC+ST & 56.7 &  54.1\\
LC+NSP+ST & 65.4 &  55.9 \\
EntLM & 68.6 &  55.2 \\
EntLM (Struct) & \textbf{69.9} &  \textbf{57.1} \\

\bottomrule

\end{tabular}
\caption{Comparison with the methods presented in \cite{huang-etal-2021-shot}. LC is linear classifier fine-tuning method. P is prototype-based training using a nearest neighbor objective. NSP is noising supervised pre-training and ST is self-training. Notice that our method shows better results even without NSP and ST, and can also be further boosted by these two methods.}
\label{tab:comprehensive}
\vspace{-0.3cm}
\end{table}

\renewcommand\arraystretch{1.2}
\begin{table*}[hb]
\centering
\small
\begin{tabular}{lll}
\hline
\hline
\multirow{2}{*}{\textbf{Datasets}}  & \multirow{2}{*}{\textbf{Label words (Data\&LM+Virtual Search)}} \\  \\
\hline

\multirow{3}{*}{\textbf{CoNLL'03}}       & \{"I-PER": ["Michael", "John", "David", "Thomas", "Martin", "Paul"], "I-ORG": ["Corp", "Inc", \\&"Commission", "Union", "Bank", "Party"], "I-LOC": ["England", "Germany", "Australia", "France",\\& "Russia", "Italy"], "I-MISC": ["Palestinians", "Russian", "Chinese", "Russians", "English", "Olympic"]\}    \\
\midrule
\multirow{8}{*}{\textbf{OntoNotes*}}   & \{"I-EVENT": ["War", "Games", "Katrina", "Year", "Hurricane", "II"], "I-FAC": ["Airport", "Bridge", \\&"Base", "Memorial", "Canal", "Guantanamo"], "I-GPE": [    "US", "China", "United", "Beijing", \\&"Israel", "Taiwan"], "I-LANGUAGE": ["Mandarin", "Streetspeak", "Romance", "Ogilvyspeak",\\& "Pentagonese", "Pilipino"],     "I-LAW": ["Chapter", "Constitution", "Code", "Amendment", "Protocol", \\&"RICO"], "I-LOC": ["Middle", "River", "Sea", "Ocean", "Mars", "Mountains"], "I-NORP    ": ["Chinese",\\& "Israeli", "Palestinians", "American", "Japanese", "Palestinian"], "I-ORG": ["National", "Corp", "News",\\& "Inc", "Senate", "Court"], "I-PERSON": ["John", "David", "Peter", "Michael", "Robert", "James"],\\& "I-PRODUCT": ["USS", "Discovery", "Cole", "Atlantis", "Coke", "Galileo"],\\& "I-WORK\_OF\_ART"    : ["Prize", "Nobel", "Late", "Morning", "PhD", "Edition"]\} \\
\midrule

\multirow{7}{*}{\textbf{MIT-Movie}}      & \{"I-ACTOR": ["al", "jack", "bill", "pat", "der", "mac"], "I-CHARACTER": ["solo"], \\&"I-DIRECTOR": ["de", "del",  "stone", "marks", "bell", "dick"], "I-GENRE    ": ["fantasy", "adventure",\\& "romance", "comedy", "action", "thriller"], "I-PLOT": ["murder", "death", "vampires", "aliens",\\& "zombies", "suicide"], "I-RATING": ["13"],  "I-RATINGS\_AVERAGE": ["very", "nine", "well",\\& "highly", "really", "popular"], "I-REVIEW": ["comments", "regarded", "opinions", "positive"],  \\&   "I-SONG": ["heart", "favourite", "loves"],  "I-TITLE": ["man", "woman", "night", "story", "men", \\&"dark"], "I-TRAILER": ["trailers", "trailer", "preview",     "glimpse", "clips"],\\& "I-YEAR": ["last", "past", "years", "decades", "ten", "three"]\}   \\

\hline
\hline
\end{tabular}
\caption{Label words obtained by Data\&LM+Virtual Search method. The number of label words for each class might be less than $k=6$ if the words cannot meet the conflict threshold $Th=0.6$.}
\label{tab:label_word_example}
\vspace{-0.2cm}
\end{table*}

\newpage

\begin{algorithm}\label{support}

\caption{Few-shot Sampling}
\small
\label{algo:support}
\begin{algorithmic}[1]
\REQUIRE \# of shot $K$, labeled training set $\mathbf{D}$ with a label set $\mathcal{Y}$.

\STATE $S \gets \phi $ \textcolor[rgb]{0,0.3,0.9}{\textit{// Initialize the support set}}
\FOR{each class $i \in \mathcal{Y}$}
\STATE $ \text{Count}[i] \gets 0 $
 \textcolor[rgb]{0,0.3,0.9}{// \textit{Initialize the counts of each entity class}}
\ENDFOR

\STATE {Shuffle} $\mathbf{D}$
\FOR{$(X,Y) \in \mathbf{D}$}
\STATE $\text{Add} \gets \text{True}$
\FOR{$i \in \mathcal{Y}$ }
\STATE Calculate Temp\_count[i]  
\textcolor[rgb]{0,0.3,0.9}{// \textit{Calculate the mention number of class $i$ in $(X,Y)$}}
\IF{$\text{Count}[i] + \text{Temp\_count}[i] > K$}
\STATE $\text{Add} \gets \text{False}$ 
\textcolor[rgb]{0,0.3,0.9}{// \textit{Skip current sample that violates the $K$-shot rule}}
\ENDIF
\ENDFOR
\IF{Add is True}
\STATE $S \gets S \cup \{ ({X},{Y}) \}$
\STATE Update $\{ \text{Count}[i] \gets \text{Count}[i] + \text{Temp\_count}[i] \}$ $\forall i \in \mathcal{Y}$
\ENDIF
\IF{$\text{Count}[i] == K, \forall i \in \mathcal{Y}$}
\STATE break
\textcolor[rgb]{0,0.3,0.9}{// \textit{Finish sampling}}
\ENDIF

\ENDFOR

\RETURN $\mathcal{S}$
\end{algorithmic}
\end{algorithm}


\end{document}